\documentclass[conference]{IEEEtran}

\newcommand{\ignore}[1]{}
\usepackage[pass]{geometry}
\usepackage{fancyhdr}
\usepackage[normalem]{ulem}
\usepackage[hyphens]{url}

\usepackage{amsmath}
\usepackage{graphics}
\usepackage[font=md]{subcaption}
\usepackage[font=bf]{caption}
\usepackage{listings}
\usepackage{booktabs}
\usepackage{adjustbox}
\usepackage{afterpage}
\usepackage{mathtools} 
\usepackage{amsfonts}
\usepackage{amssymb}
\usepackage{tabulary}
\usepackage{multirow}
\usepackage{soul}
\usepackage[protrusion=true,tracking=true,kerning=true,expansion,final]{microtype}
\usepackage[pagebackref]{hyperref}

\newcommand{\ra}[1]{\renewcommand{\arraystretch}{#1}}

\begin{document}
	\title{Streaming Architecture for Large-Scale Quantized Neural Networks on an FPGA-Based Dataflow Platform}

	\author{\IEEEauthorblockN{Chaim Baskin, Evgenii Zheltonozhskii, \\
			Alex M. Bronstein and Avi Mendelson}
		\IEEEauthorblockA{Department of Computer Science\\
			Technion -- Israel Institute of Technology\\
			Haifa, Israel\\
			\href{mailto:chaimbaskin@cs.technion.ac.il}{chaimbaskin@cs.technion.ac.il},
			\href{mailto:evgeniizh@campus.technion.ac.il}{evgeniizh@campus.technion.ac.il}, \\  
			\href{bron@cs.technion.ac.il}{bron@cs.technion.ac.il},
			\href{mailto:avi.mendelson@tce.technion.ac.il}{avi.mendelson@tce.technion.ac.il}}
		\and
		\IEEEauthorblockN{Natan Liss}
		\IEEEauthorblockA{\\Department of Electrical  Engineering\\
			Technion -- Israel Institute of Technology\\
			Haifa, Israel\\
			\href{mailto:lissnatan@campus.technion.ac.il}{lissnatan@campus.technion.ac.il}}}
	
	\maketitle
	
	\begin{abstract}
		Deep neural networks (DNNs) are used by different applications that are executed on a range of computer architectures, from IoT devices to supercomputers. The footprint of these networks is huge as well as their computational and communication needs. In order to ease the pressure on resources, research indicates that in many cases a low precision representation (1-2 bit per parameter) of weights and other parameters can achieve similar accuracy while requiring less resources. Using quantized values enables the use of FPGAs to run NNs, since FPGAs are well fitted to these primitives; e.g., FPGAs provide efficient support for bitwise operations and can work with arbitrary-precision representation of numbers.
		
		This paper presents a new streaming architecture for running QNNs on FPGAs. The proposed architecture scales out better than alternatives, allowing us to take advantage of systems with multiple FPGAs. We also included support for skip connections, that are used in state-of-the art NNs, and shown that our architecture allows to add those connections almost for free. All this allowed us to implement an 18-layer ResNet for $224\times224$ images classification, achieving $57.5\%$ top-1 accuracy.
		
		In addition, we implemented a full-sized quantized AlexNet. In contrast to previous works, we use 2-bit activations instead of 1-bit ones, which improves AlexNet's top-1 accuracy from $41.8\%$ to $51.03\%$ for the ImageNet classification. Both AlexNet and ResNet can handle 1000-class real-time classification on an FPGA. 
		
		Our implementation of ResNet-18 consumes $5\times$ less power and is $4\times$ slower for ImageNet, when compared to the same NN on the latest Nvidia GPUs. Smaller NNs, that fit a single FPGA, are running faster then on GPUs on small ($32\times32$) inputs, while consuming up to $20\times$ less energy and power.
		
	\end{abstract}
	

	\IEEEpeerreviewmaketitle

	\section{Introduction}
	A neural network (NN) \cite{Caudill:1989:NNP:69737.69743} \cite{Patterson:1998:ANN:521611} is a computational model inspired by the way we believe our brain operates: the data that comes from our sensors, e.g., eyes, is processed by multiple simple computational units called neurons. The neurons are interconnected through a complex network of connections (axons), and after several transformations, the input is translated into a conclusion such as ``there is a chair in the picture.'' Similarly, artificial NNs use vast amounts of simple computational elements that are organized in interconnected layers. Modern NNs usually have multiple layers (sometimes 1000 \cite{he2016deep} or more) and thus are called deep neural networks (DNNs). These networks are widely used in image processing, medicine, autonomous driving, translation and other fields.
	
	In order to better interpret local features of multidimensional inputs such as images, convolutional neural networks (CNNs) are commonly used. This type of NNs has been shown to be efficient in image-related problems such as classification or scene parsing. To achieve these results, CNNs need many parameters (over 100M parameters reported in \cite{DBLP:journals/corr/SimonyanZ14a}) and require huge amounts of computational resources and memory. As a result, expensive and power hungry computers are needed to efficiently process these networks, which has led researchers to seek ways to reduce the computational, memory, and bandwidth requirements \cite{DBLP:journals/corr/CourbariauxBD14} \cite{chen2015compressing} \cite{DBLP:journals/corr/HanMD15} \cite{gupta2015deep} \cite{DBLP:journals/corr/IandolaMAHDK16}.
	
	Using binarized neural networks (BNNs) \cite{golea1992learning} \cite{DBLP:journals/corr/KimS16} \cite{courbariaux2015binaryconnect} is one proposed solution to the problem. In BNNs, each parameter is represented by only one bit, which saves memory, communication time and energy, and enables the use of bitwise operations, which are simpler and faster than multiplications. For this reason, FPGAs seem to be the most appropriate architecture for BNN execution. Programming FPGAs, however, is non-trivial, especially in comparison to modern scripting languages that are being used for NN development. In order to simplify development, major FPGA manufacturers have invested heavily in high-level synthesis tools that can translate a program written in a high level language such as OpenSPL\cite{Becker2016} and C-to-VHDL (presented as part of Vivado HLS \cite{winterstein2013high}), or frameworks such as OpenCL \cite{munshi2009opencl} \cite{singh2011implementing}. Today, HLS-based tools provide a decent trade-off between resource utilization, compared to custom-written HDL code, and development time. 
	
	In this paper, we focus on architectural and optimization techniques for implementing QNNs on FPGAs using high level programming languages. The main objective of this work is to investigate architectural features of reduced-precision NNs without focusing on low-level optimizations, and accordingly we used an HLS-based platform to model our architecture. We propose a streaming model based on functional decomposition of the computations, which are embedded in data flow engines (DFEs) based on FPGAs. For this purpose, we used the OpenSPL programming environment and the Maxeler's hardware platform since the latter allowed us to implement the desired processor model using high level languages. 
	
	The paper indicates that QNNs scale well both on input and network sizes, showing only a minor increase in resource usage on larger inputs. In addition, our system can easily be divided into a couple of FPGAs, almost without a performance drop. All this allows us to run a full-sized ResNet-18 and AlexNet on two and three FPGAs, respectively, achieving runtime comparable with the latest GPUs, consuming less power and energy. Moreover, in contrast to previous work, we implemented multiple-bit activations, which improves accuracy of the network by up to $10\%$ \cite{DBLP:journals/corr/ZhouNZWWZ16} \cite{DBLP:journals/corr/HubaraCSEB16}.
	
	We also analyze skip connections and their impact on resource utilization and runtime, concluding that streaming architecture allows us to add skip connections for a relatively small price. 
	
	The paper is organized as follows:Section \ref{sec_fpga} explains the platform on which we built our network. Section \ref{sec_archtecture} describes our model architecture and optimizations. Section \ref{sec_eval} presents our experimental evaluation, Section \ref{sec_conclusion} presents our conclusions.

	\section{Design Methodologies for FPGA-based systems} \label{sec_fpga}
	
	FPGAs are operated at relatively low frequencies and are based on a simple execution unit that usually can operate only a couple of bits (typically less than 8). In order to achieve overall high performance and low power per operation, FPGAs rely on massively parallel operations at the chip level. 
	
	Traditionally, software for mainstream hardware is based on data decomposition: the same operations are executed in parallel on a massive amount of independent data. Another approach to achieving massively parallel operations in general---and on FPGAs in particular---is {\it functional decomposition}, also called dataflow. In this execution model, the functionality of an algorithm is decomposed into independent parallel threads and the data flows between them.
	
	Procedural languages with vector operations are good examples of the first approach, while TenserFlow and Maxeler OpenSPL are examples of the second one.
	
	\subsection {The use of pipeline parallelism for programming FPGAs}
	High level languages such as C, C++ or OpenCL are often used to program highly complicated algorithms, such as CNNs on FPGAs. Usually, a restricted version of these languages is used to simplify translation into lower-level representation by applying auto-vectorization techniques.
	
	Many systems allow the addition of specific optimizations at this lower level, e.g., an efficient implementation of the XNOR primitive. Thus, the full development path starts by implementing the entire system using a high level language, followed by gradual replacement of critical blocks with highly optimized specific synthesized blocks to optimize the system.
	\subsection {The use of functional decomposition for programing FPGAs }
	
	Designing the system based on functional decomposition starts by identifying the different functionalities the system needs to perform and determining the flow of the data between blocks performing these functions. 
	
	This approach can use the notion of dataflow in which system activities are triggered by their inputs being ready and the output buffers able to hold the results. It fits well with the concept of streaming processing where ``nodes'' are implemented as threads and data are transferred using configurable routing resources, buffered on-chip memory, and flip-flops, embedded on an FPGA.
	
	Functional decomposition has the advantage of scale-out (can easily be extended over multiple FPGAs) but needs to be designed with extra care, since a bottleneck in one of the nodes can determine the performance of the entire system.
	
	In this work, we chose to use the software environment of Maxeler's system, since it is (1) inherently built around the notion of {\it data flow engines} (DFEs) and (2) can be programmed using high level languages.  
	
\begin{figure}
		\centering
		\includegraphics[width=0.7\linewidth]{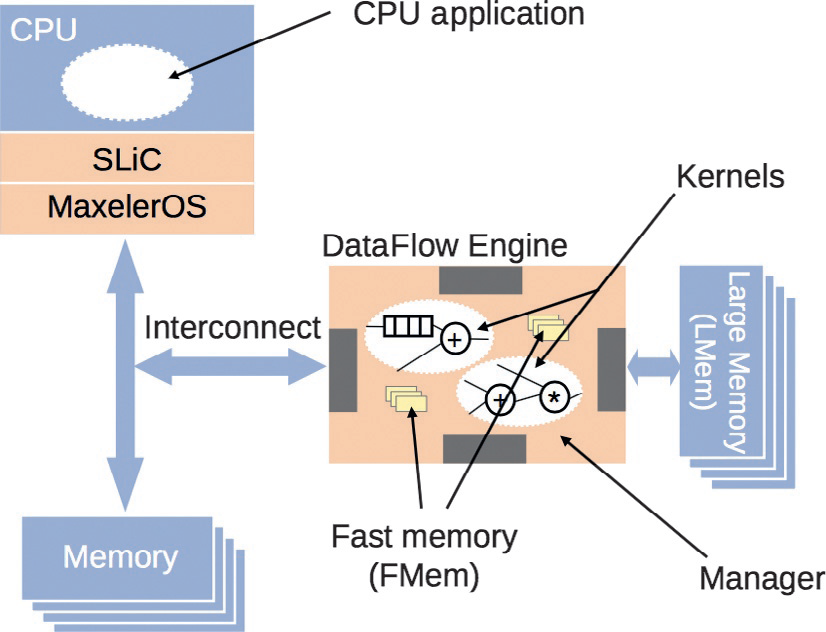}
		\caption{The architecture of a Maxeler DFE system \cite{Oriato2015105}.} 
		\label{DFE}
		\vspace{-1.5em}
	\end{figure}

	The general structure of Maxeler's environment is shown in Figure \ref{DFE}.
	
	Maxeler boards consist of multiple CPUs and multiple FPGAs. Each DFE contains a single FPGA, which interfaces with a CPU via a PCIe. Multiple DFEs are interconnected in a daisy chain topology, via a proprietary link called a MaxRing.  Figure \ref{DFE} depicts the architecture of a Maxeler dataflow processing system. 
	The Maxeler system can execute multiple kernels concurrently to support multiple streams of data both at the level of internal computations of the DFEs and between the CPU and the DFEs.
	
	Although Maxeler systems allows the attachment of a large amount of memory to each FPGA (LMem in Figure \ref{DFE}), in this work we used only the memory that is embedded in the FPGA fabric, called fast memory (FMem in Figure \ref{DFE}). FMem can store only a few megabytes of data, but the access to memory is much faster  and thus FMem can be used as a communication buffer between the DFEs.
	
	The entire system is written in high level languages: Java for the kernels and manager and C++ for the CPU code.

	\section{The proposed streaming architecture of QNNs on FPGAs} \label{sec_archtecture}
	
	This work focuses on developing a streaming architecture that uses dataflow-based functional decomposition in order to efficiently run QNNs. In this section, we describe the architecture, the optimizations and the internal structure of a system that can efficiently run different QNNs and handle inputs of any size. 
	
	\subsection {Overview of DNN architecture}
	
	We developed an architecture for regular CNNs and their main building blocks (convolutional, pooling and fully connected layers) and also for residual networks.
	Residual networks add skip connections to CNNs architecture. Skip connections forward the output of one layer to the one after adjacent one, skipping one layer. This resolves the vanishing gradient \cite{bengio1994learning} problem, thus increasing the number of layers and achieving state-of-the-art accuracy on image-related problems \cite{DBLP:journals/corr/ZagoruykoK16} \cite{inceptionv4} \cite{DBLP:journals/corr/HuangRSZKFFWSG016}.  
	We developed a hardware design for skip connections and, to analyze their performance, implemented the ResNet-18 \cite{he2016deep} network, which architecture is shown in Table \ref{ResNet}.
	
	Additionally, we implemented the AlexNet \cite{NIPS2012_4824}, since it is one of the most well-known DNNs and is often used as a basis for new techniques in DNNs such as network compression, performance improvements, and new types of layers \cite{DBLP:journals/corr/HanMD15} \cite{dicecco2016caffeinated} \cite{rastegari2016xnor} \cite{DBLP:journals/corr/IandolaMAHDK16} \cite{DBLP:journals/corr/MiyashitaLM16} \cite{DBLP:journals/corr/HubaraCSEB16} \cite{DBLP:journals/corr/AydonatOCLC17}.
	The network consists of eight layers: the first five are convolutions intermediated with pooling layers, and the remaining three are fully connected.
	The output of the last layer is fed to a 1000-way softmax, which produces a distribution over the 1000 class labels.
	\newcommand{\blocka}[2]{\multirow{3}{*}{\(\left[\begin{array}{c}\text{3$\times$3, #1}\\[-.1em] \text{3$\times$3, #1} \end{array}\right]\)$\times$#2}
	}
	\newcommand{\blockb}[3]{\multirow{3}{*}{\(\left[\begin{array}{c}\text{1$\times$1, #2}\\[-.1em] \text{3$\times$3, #2}\\[-.1em] \text{1$\times$1, #1}\end{array}\right]\)$\times$#3}
	}
	\renewcommand\arraystretch{1.1}
	\setlength{\tabcolsep}{3pt}
	\begin{table}[t]
		\begin{center}
			\resizebox{0.7\linewidth}{!}{
				\begin{tabular}{c|c|c}
					\hline
					Layer name & Output size & Layer parameters \\
					\hline
					conv1 & 112$\times$112 & 7$\times$7, 64, stride 2\\
					\hline
					\multirow{4}{*}{conv2\_x} & \multirow{4}{*}{56$\times$56} & 3$\times$3 max pool, stride 2 \\\cline{3-3}
					&  & \blocka{64}{2}  \\
					&  &  \\
					&  &  \\
					\hline
					\multirow{3}{*}{conv3\_x} &  \multirow{3}{*}{28$\times$28}  & \blocka{128}{2} \\
					&  &  \\
					&  &   \\
					\hline
					\multirow{3}{*}{conv4\_x} & \multirow{3}{*}{14$\times$14}  & \blocka{256}{2}  \\
					&   \\
					&   \\
					\hline
					\multirow{3}{*}{conv5\_x} & \multirow{3}{*}{7$\times$7}  & \blocka{512}{2}  \\
					&  &  \\
					&  &  \\
					\hline
					& 1$\times$1  & average pool, 1000-d fc, softmax\\
					\hline
				\end{tabular}
			}
		\end{center}
		\vspace{-.5em}
		\caption{ResNet-18 archtecture from \cite{he2016deep}. Brackets contain one block, and each block is stacked twice. conv3\_1, conv4\_1 and conv5\_1 have a stride of 2 to perform downsampling.
		}
		\label{ResNet}
		\vspace{-.5em}
	\end{table}
	
	\subsection {Hardware implementation overview}
	
	CNN models used in our evaluations (ResNet and AlexNet) are based on the work of Hubara et al.\ \cite{DBLP:journals/corr/HubaraCSEB16}. 
	We chose to use $1$-bit weights and $2$-bit activation function outputs. According to Hubara's evaluations, this set of parameters is a satisfactory compromise between memory requirements and model accuracy. 
	
	All the pre-trained weights and normalization parameters are stored on the CPU side, while all the computations required for the inference are performed on the DFE side. 
	In order to fully utilize the DFE's spatial computation capabilities, we chose a streaming architecture in which the output of each layer is fed to the input of the next one as shown in Figure \ref{conv}. Unlike a traditional approach, in which the computation of the current layer starts once the previous one has finished, streaming architecture allows the current layer to begin its output calculation once enough data has been accumulated in its internal buffer. Moreover, in streaming architecture there is no need to store each layer's intermediate results in off-chip memory, since they are immediately passed down the stream.  
	
	The input to each kernel, which represents an NN layer, is a stream of pixels stored in an internal buffer (Shift Register in Figure \ref{conv}). As soon as all the data required (shown as a stack of pixels in Current Window  in Figure \ref{conv}) for the calculation of the particular output pixel is present, the pixel is calculated and passed to the next layer. It means we can treat other layers as a black box that receives or provides pixels. This approach simplifies integration of layers and building of complicated networks. Since each layer is represented in the DFE Manager by a single function call, the  building of the network is similar to the process of building in high level frameworks.
	
	Each kernel starts the computation as soon as the previous one provides output to it. Due to this computation overlap, the latency is pretty small, and after the initiation interval, computations are performed by all layers simultaneously.   
	Additionally, due to the model's compact size, all NN parameters are kept in on-chip memory, eliminating the need to use slower off-chip memory. Further subsections describe the hardware design of each QNN component.
	
	\begin{figure}
		\centering
		\includegraphics[width=\linewidth]{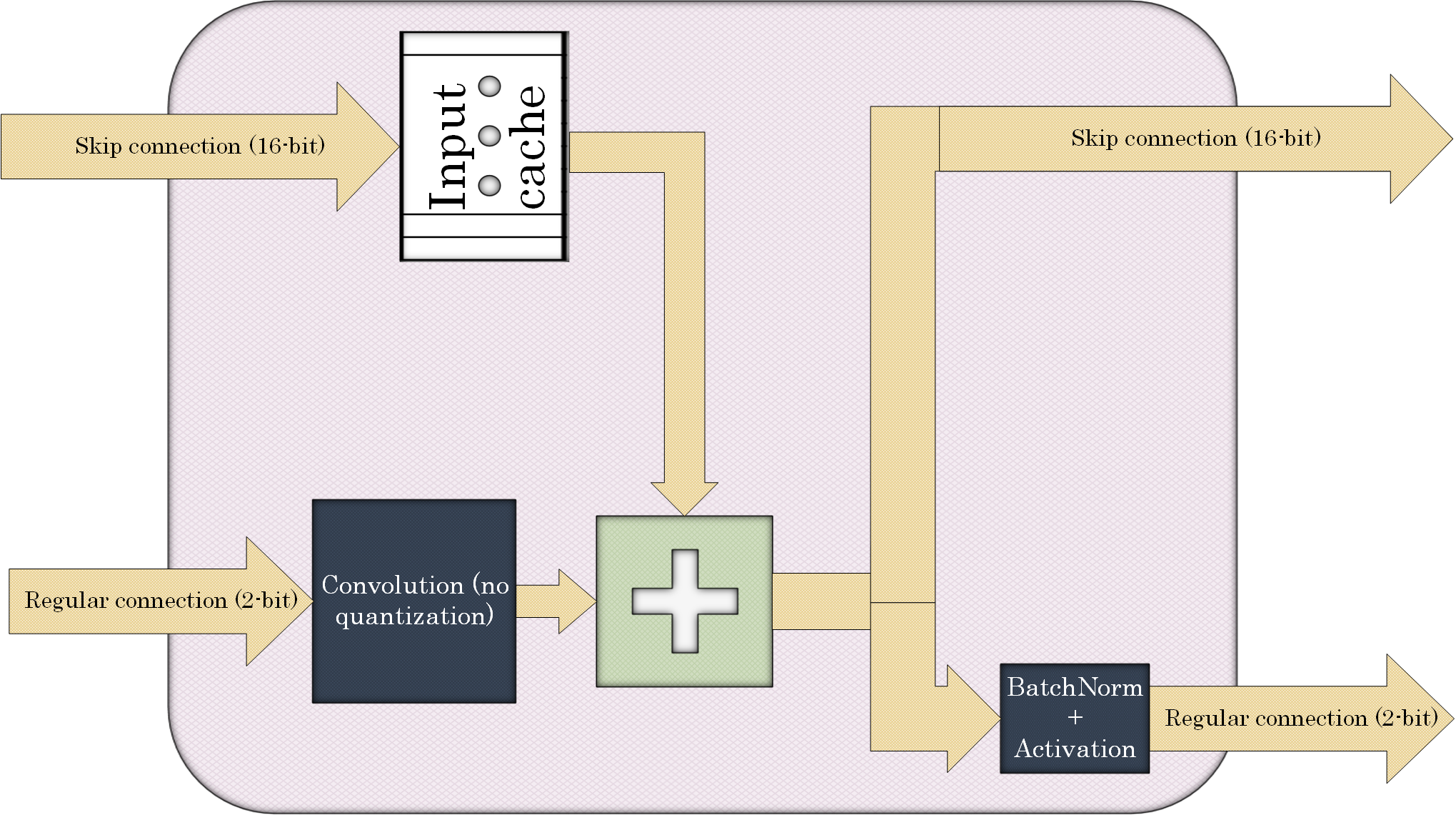}
		\vspace{-1.2em}
		\caption{Skip connection processing. Convolution output is summed with input from skip connection and passed both to regular and skip connection.}
		\label{skip}
		\vspace{-1em}
	\end{figure}
	
	\subsubsection {Convolution} \label{conv_subs}
	\begin{figure}
		\centering
		\includegraphics[width=\linewidth]{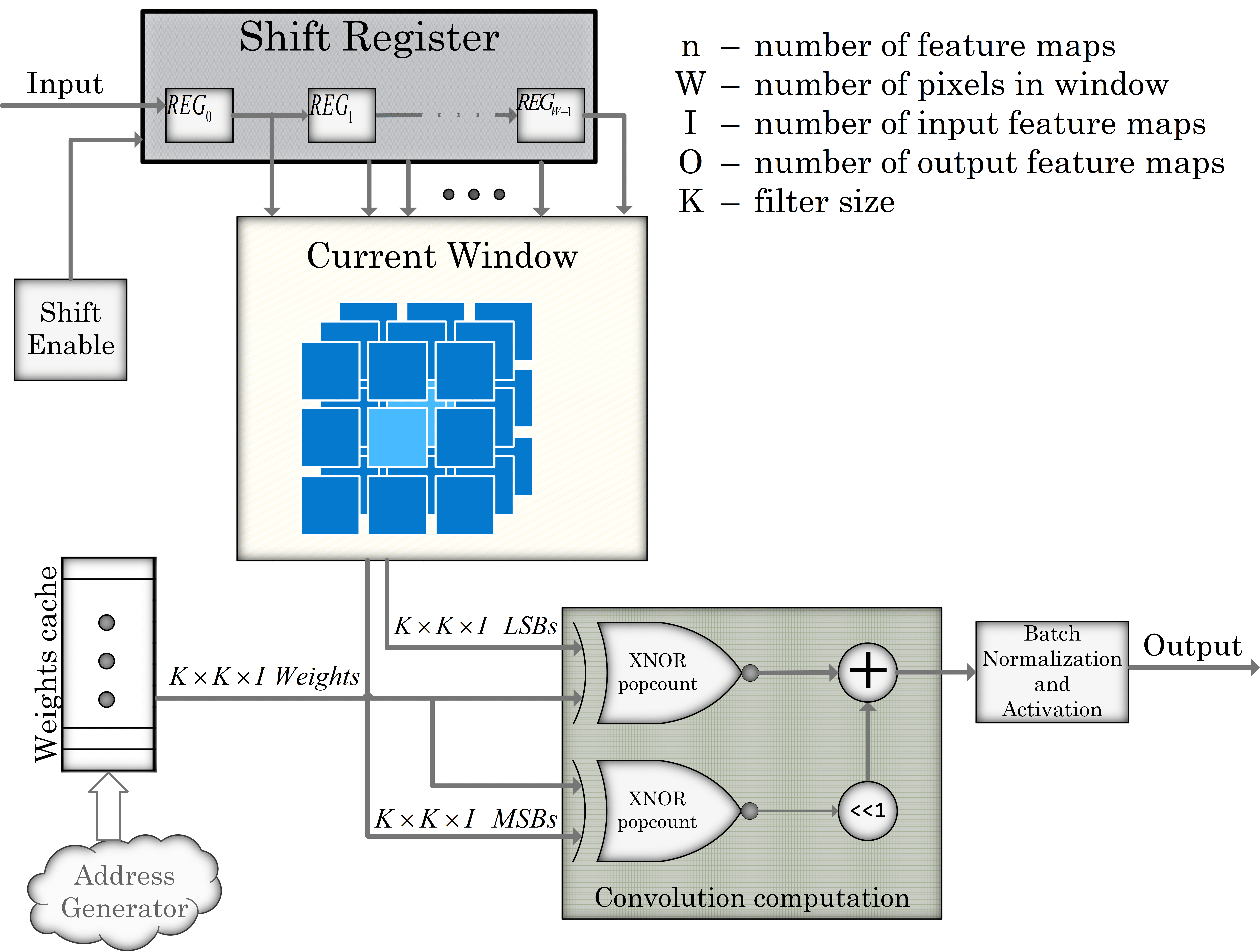}
		\vspace{-1.2em}
		\caption{Convolution kernel dataflow. The partial case where input is 2-bit is shown.}
		\label{conv}
		\vspace{-1em}
	\end{figure} 
	
	The execution of the convolution kernel (Figure \ref{conv})  starts with inputs for weights, BatchNorm parameters, and feature maps. Pixels that are currently processed are stored in shift registers, while binarized weights and BatchNorm parameters are stored in the FPGA's internal memory caches. We replaced element-wise matrix multiplication of feature maps and their corresponding weights with the XNOR-popcount algorithm, followed by BatchNorm and activation functions.
	
	The inference begins with fetching parameters: weights, biases and BatchNorm parameters. After all the parameters have been fetched, we start to input the feature maps. Every time there is enough data in the internal shift register, the kernel halts the input and calculates one output pixel per clock cycle, until all the filters are applied at this position (i.e., same (X,Y) coordinates in all feature maps). 
	There are positions that do not produce any output; for example, the borders of the input feature map and, in the case of strided convolution, all pixels between two valid filter positions. This is especially important in the first layer, where, given the stride $S=4$, we acquire around 13$\times$ speedup.
	
	If the image is padded, then, when the kernel is processing padding pixels, it stops the input stream and inputs padding values into the buffer instead. The only available values for BNNs are $-1$ and $1$, meaning zero-padding is not possible, and $-1$ padding was used instead.
	\paragraph{Weights and BatchNorm coefficient storage}
	All the weights received by the FPGA are represented as 32-bit floating point numbers. Before storing these parameters in the internal memory cache, we transformed them into a 1-bit representation, using the $Sign$ function, as described earlier.
	
	For the filter dimensions $K\times K\times I$, where $K$ is the size of the filter and $I$ is the number of input feature maps, there are $K\times K\times I \times O$ weights at this layer, where $O$ is the number of output feature maps. In order to calculate one output pixel, we need to access $K\times K\times I$ weights simultaneously. Therefore, each address of the cache stores $K\times K\times I$ weights and the cache has $O$ entries.
	
	Since BRAMs have a limited number of predefined width/depth configurations, there is no way to avoid overhead while storing weights. In our FPGA, the minimal depth of a BRAM is 512, while the maximal number of weight cache entries is 384. A BRAM can allow only one access per clock, which means that at least $25\%$ of each BRAM used for weights cache is wasted.
	
	The amount of memory required for normalization parameter storage is relatively small. We need to store $2\times O$ normalization parameters for each layer in its cache. Both  parameters are represented as 32-bit integers and stored as a single 64-bit number. This means that each layer's normalization parameter cache has $O$ entries of 64 bits each.   
	
	The weights and normalization parameters enter each layer in depth-first order, similarly to the feature maps. They are loaded into their dedicated caches only once, before inference of images starts, and then used repeatedly during inference. 
	
\begin{figure}	
		\begin{subfigure}{\linewidth}
			\includegraphics[width=\linewidth]{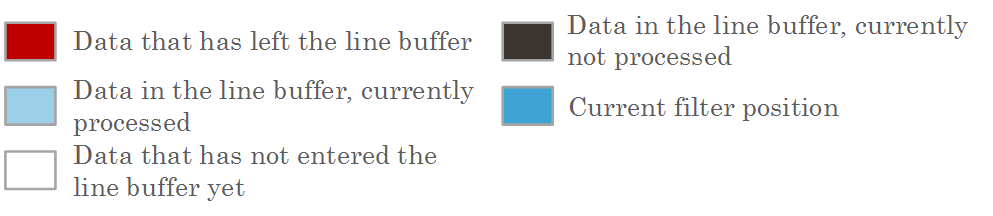}
		\end{subfigure}
		
		\begin{subfigure}[b]{0.4565\linewidth}
			\includegraphics[width=\linewidth]{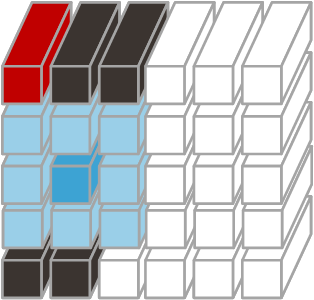}
			\caption{Depth-first scan} \label{conv_scan_depth}
		\end{subfigure}
		\hspace*{\fill} 
		\begin{subfigure}[b]{0.5017\linewidth}
			\includegraphics[width=\linewidth]{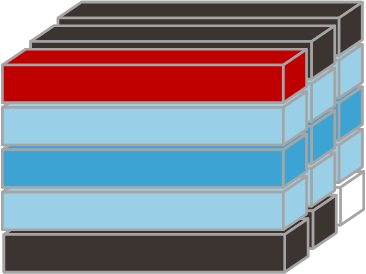}
			\caption{Width-first scan} \label{conv_scan_width}
		\end{subfigure}
		\vspace{-0.5em}
		\caption{Comparison of depth-first and width-first scans}
		\label{conv_scan} 
		\vspace{-1.2em}
	\end{figure}

	\paragraph{Feature map buffering}
	Let us define an input tensor of size $H\times W\times I$, and a filter tensor of size $K\times K\times I\times O$. In order to calculate the first output pixel, we can choose two possible options to scan the input pixels, as shown in Figure \ref{conv_scan}. The necessary buffer size for Figure \ref{conv_scan_depth} is $I\times H\times (K-1) + I \times K$, and the size for Figure \ref{conv_scan_width} is $H\times W\times (I-1) + H\times (K-1)+ K$, which means memory requirements per height dimension for the two methods are $\Theta(IK)$ and $\Theta(IW+K)$, respectively. Since $W > K$ (sometimes an order of magnitude bigger), scanning to depth guarantees a smaller buffer.     
	This means that in order to minimize the number of flip flops used for feature map buffering, all images should be streamed to the FPGA pixel by pixel and not channel by channel.
	
	\subsubsection {Pooling}
	The pooling kernel is built similarly to the convolutional one. Since the pooling has no parameters, output pixels are calculated as soon as enough data is accumulated inside the internal buffers. In addition, since each output pixel depends only on its own feature map, we do not need to wait until input is finished, but can produce output at the same clock cycle at which the input is received.
	In our implementation, max pooling is used in all cases, except for the last pooling in ResNet-18.

	\subsubsection {Batch normalization and activation function}
	As was shown in FINN \cite{Umuroglu:2017:FFF:3020078.3021744}, BatchNorm and one-bit activation can be replaced by a threshold function. We extend this idea to multiple-bit activations, performing BatchNorm and $n$-bit activation using only two additional parameters with an $n$-input comparator and a $2^n\to 1$ multiplexer. 
	
	Using the notation of \cite{Umuroglu:2017:FFF:3020078.3021744}, we denote pre-activation output of neuron $k$ as $a_k$, and BatchNorm parameters as $\Theta_k = \left( \gamma_k, \mu_k, i_k, B_k \right)$. Then BatchNorm is calculated as $BatchNorm\left(a_k,\Theta_k\right) = \gamma_k \cdot \left(a_k - \mu_k\right) \cdot i_k + B_k $. 
	The $n$-bit uniform activation (quantization) divides the range of inputs into $2^n$ equally-sized ranges. Each range is mapped to a single output value of the activation function.
	Denote the size of each range as $d$. Given the mean $\mu$ and $d$, we can calculate the endpoints of all ranges. Thus, to acquire an output of the normalization and activation function combination for a pre-normalized value (i.e.,  which range it belongs to), it is enough to have a value of one of the endpoints and the size of the range.
	To this end, we first solve $BatchNorm\left(\tau_k,\Theta_k\right) = 0$, acquiring $\tau_k = \mu_k - B_k / \left(\gamma_k \cdot i_k\right)$. Next, by solving $BatchNorm\left(t_k,\Theta_k\right) = \alpha \cdot d$, we acquire $t_k = \mu_k + \left(\alpha \cdot d - B_k\right)/\left(\gamma_k \cdot i_k\right) = \tau_k + \alpha  \cdot \left[d/\left(\gamma_k \cdot i_k\right) \right] $. Therefore, to calculate all endpoints, it is enough to have $\tau_k$ and $d/\left(\gamma_k \cdot i_k\right)$. Finally, we perform a binary search on the ranges to determine in which range $a_k$ falls. 
	
	\subsubsection {Fully connected layer}
	As shown by Springenberg et al.\ \cite{DBLP:journals/corr/SpringenbergDBR14}, the traditional architecture of convolutional layers followed by FC layers can be replaced by an all-convolutional network (i.e., an NN that consists only of convolutional and pooling layers) where FC layers are represented as 1-by-1 convolutions. The specifics of fully connected layers---large amounts of weights and small amounts of neurons---influences resource utilization: more BRAMs, but less LUTs and FFs are used.
	\subsubsection{Skip connections}
	Skip connections are implemented as a part of residual network building block, which contains two convolutional layers and additional infrastructure to manage a skip connection, namely, a buffer and an adder. As shown in Figure \ref{skip}, the block receives two inputs: one via a skip connection and one via a regular.
	
	The data passed in skip connections are 16-bit integers, which accumulate non-quantized outputs of convolutions. The whole block works as follows: the regular connection input, which is, as described earlier, 2-bits wide, enters a convolution block (\ref{conv_subs}). At this stage, BatchNorm and activation are not applied. The convolution output is summed with input from the skip connection and the result is split into two paths. The first one is a skip connection, where data is sent as is. The second one goes through BatchNorm and activation, and then is streamed to the next (regular) convolution. The output of the next convolution together with the skip connection are inputs of the next ``residual block''.
	
	In order to sum the skip connection data and the corresponding convolution result, skip connection inputs are buffered to compensate for delay created by the intermediate convolutional layer in a ``regular'' path. The required buffer is exactly same size as the buffer in a convolutional layer. This is not accidental. Using previous notation, taking padding and the fact that $I=O$ into account, $I\times \left[H\times (K-1) + K\right]$ inputs in the first convolution produce $I\times \left[H\times \frac{K-1}{2} + K\right]$ inputs in the second convolution. This, together with padding, is exactly the amount of data needed to create one output pixel.
	
	From the hardware perspective, the addition of a skip connection requires a minimal amount of resources---one adder and the buffer as described earlier. The skip buffer is needed to compensate for the delay and never creates delays by itself. This means that generally, the overhead of the addition of a skip connection is negligible.
	
	\subsubsection{Multi-DFE implementation}
	Since our architecture comprises independent kernels and the Maxeler platform allows data to directly flow from DFE to DFE, the workload can be divided into multiple DFEs with very small performance degradation if the design cannot fit one DFE. Since each pixel is represented by 2 bits, the required bandwidth of the DFE-to-DFE link, for a 105 MHz fabric clock, is 210 Mbps. According to the Maxeler specifications, this link can be set to rates of up to several Gbps, which is more than enough for our purposes.
	
	\section{Evaluation} \label{sec_eval}
	We conducted our experiments on different platforms, including last-generation Nvidia GPUs and Intel FPGAs. As an FPGA-based system, we used Maxeler's MPC-X node that provides 8 MAX4 (Maia) DFEs interconnected by a dedicated MaxRing connection. Each DFE contained an Intel Stratix V 5SGSD8 FPGA. GPUs used as baseline were Nvidia's TeslaP100-12GB and Geforce GTX1080. Table \ref{hard_spec} shows the hardware specifications of the GPUs and FPGAs used for evaluation.  
	\begin{table}
		\centering
		\ra{1.3}
		
		\begin{subfigure}{0.45\linewidth}
			
			\begin{adjustbox}{width=\linewidth}
				\begin{tabular}{@{}lrr@{}} \toprule
					&P100 &GTX1080 \\  \midrule
					Architecture&Pascal&Pascal\\
					CUDA cores&3584&2560\\
					Core clock  & 1480 MHz & 1733 MHz\\ 
					\bottomrule
				\end{tabular}
			\end{adjustbox}
			\caption{GPUs specification}
			\label{gpu_spec}
		\end{subfigure}
		\hspace*{\fill} 
		\begin{subfigure}{0.45\linewidth}
			
			\begin{adjustbox}{width=\linewidth}
				\begin{tabular}{@{}lrr@{}} \toprule
					&Stratix V 5SGSD8\\ \midrule
					ALMs&262400\\
					M20K Blocks
					&2567\\
					FFs&1050K\\
					\bottomrule
				\end{tabular}
			\end{adjustbox}
			\label{fpga_spec}
			\caption{FPGA Specification}
		\end{subfigure}
		\vspace{-.5em}
		\caption{Hardware spec}
		\label{hard_spec}
		\vspace{-1em}
	\end{table}

	We measured performance, power consumption and resource utilization for FPGA implementation for three common datasets: CIFAR-10\cite{krizhevsky2009learning}, ImageNet\cite{ILSVRC15}, and STL-10 \cite{coates2011analysis}. For our evaluation, we implemented ResNet-18, AlexNet and a VGG-like CNN, based on one proposed by Umuroglu et al.\ \cite{Umuroglu:2017:FFF:3020078.3021744}, on both DFEs and GPUs. The VGG-like CNN consisted of three blocks of two convolutions and one pooling layer, and three FC layers at the end. First, the CNNs were trained for the above-mentioned datasets, using GPUs to obtain the network parameters, i.e., weights and normalization values. These parameters were then loaded onto the DFEs prior to the inference process. 
	\subsection{Methodology}\label{perf_eval}
	\paragraph*{Runtime measurements} 
	We compared the execution time of our hardware design to the execution time of two different GPUs, using the code provided by Itay Hubara on the Theano framework. Baseline timings were obtained by running 50,000 pictures through the network and taking the average. For the DFE, we similarly ran our implementation 50,000 times and took the average.
	To achieve the fastest possible execution time for the GPU, we used the latest version of Theano, which has been configured to use the NVIDIA cuDNN library. 
	\paragraph*{FPGA-based platform details}
	The kernels written in Java code were translated into VHDL by Maxeler's MaxCompiler and thereafter synthesized by Quartus to run on an FPGA. MaxCompiler generates code in MaxJ, which is a low-level Java-based hardware description language. Eventually, a bit-stream is created and downloaded to the DFE at runtime. We obtained the resource utilization, timing analysis and power estimation of the board housing the DFEs. Board power measurements were obtained using Maxeler's library called from host code.

	\subsection {Results}
	\label{results}
	This section characterizes our proposed streaming solution in terms of power, performance and scalability. We compare these parameters using different input sizes, up to $224\times224$. We also compare our results with the results of the same network running on a GPU using the Theano framework and the results claimed by Umuroglu et al.\ \cite{Umuroglu:2017:FFF:3020078.3021744} for input sizes as described in their paper. 
	
	\subsubsection{Performance against GPU-based implementation}
	\label{gpu_per}
	We compared our implementation with QNN using Hubara's code \cite{DBLP:journals/corr/HubaraCSEB16} running on two different GPU-based systems. For comparison, we chose three datasets with different input sizes ranging from $32\times32$ to $224\times224$. To show performance variation for different input sizes, we also used STL-10 resized to $144 \times 144$. For the full-sized ImageNet dataset of size $224\times224$, we used the ResNet-18 and AlexNet model, while for other inputs we used a VGG-like CNN, based on one used by Umuroglu et al., as its topology is more suitable for the above-mentioned datasets. 
	
\begin{figure}
		\centering
		
		\includegraphics[width=\linewidth]{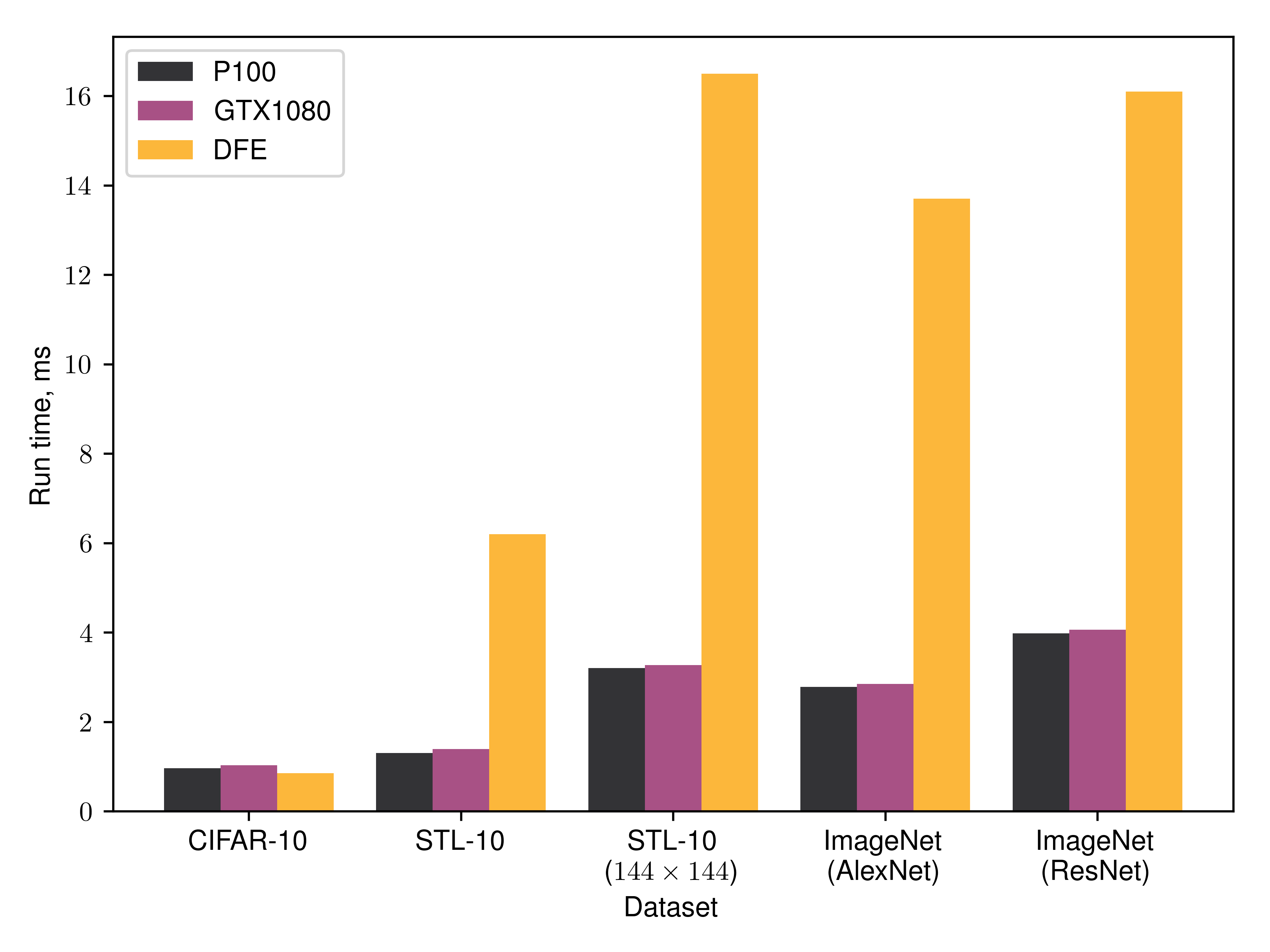}
		\vspace{-2em}
		\caption{Runtime comparison of our architecture against GPU (ms)}
		\label{time_bars}
	\end{figure}

\begin{figure}
		\centering
		
		\includegraphics[width=\linewidth]{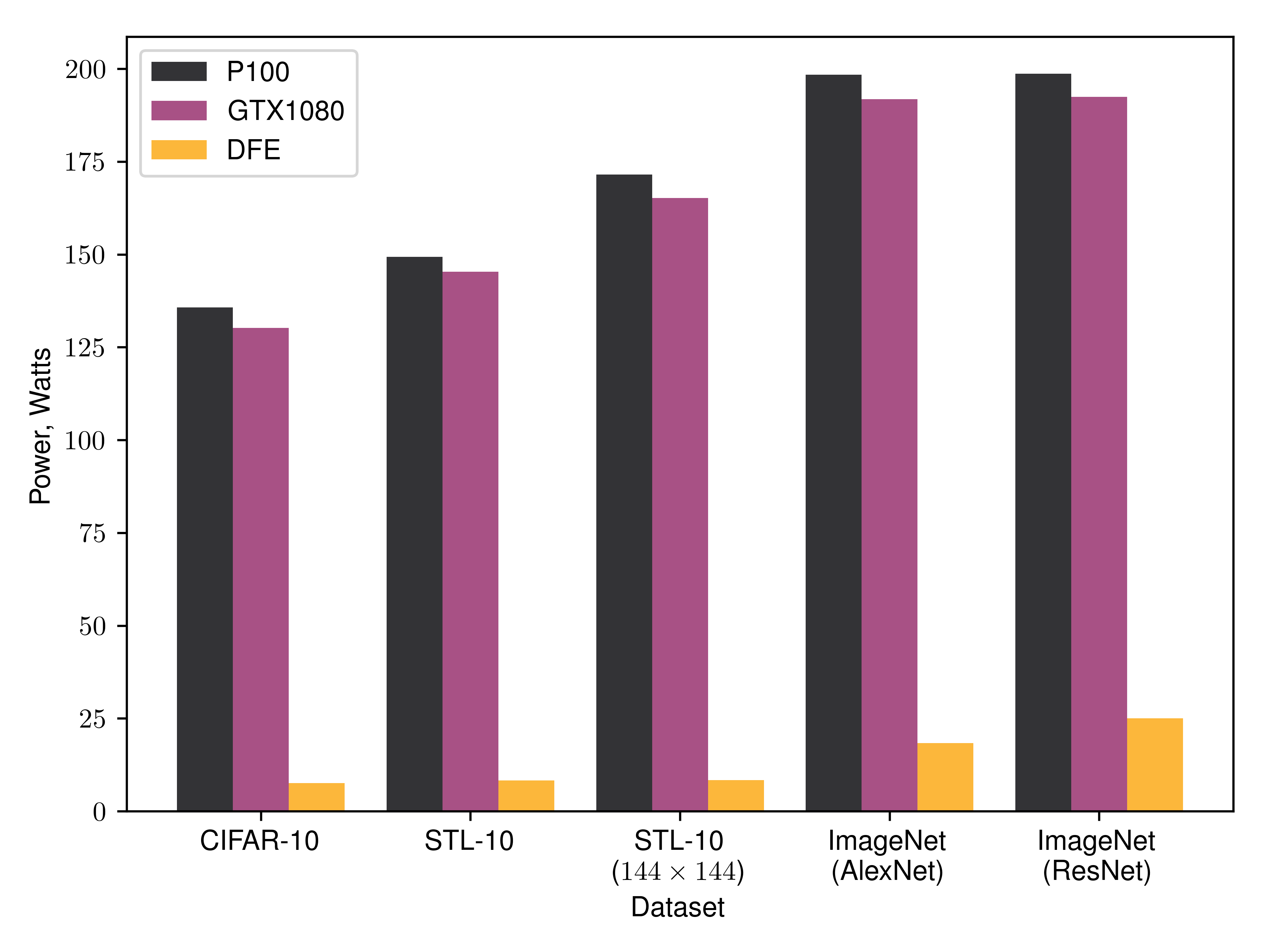}
		\vspace{-2em}
		\caption{Power consumption comparison of FPGA- and GPU-based systems (Watt)}
		\label{power_bars}
		\vspace{-1.2em}
	\end{figure}

\begin{figure}
		\centering
		
		\includegraphics[width=\linewidth]{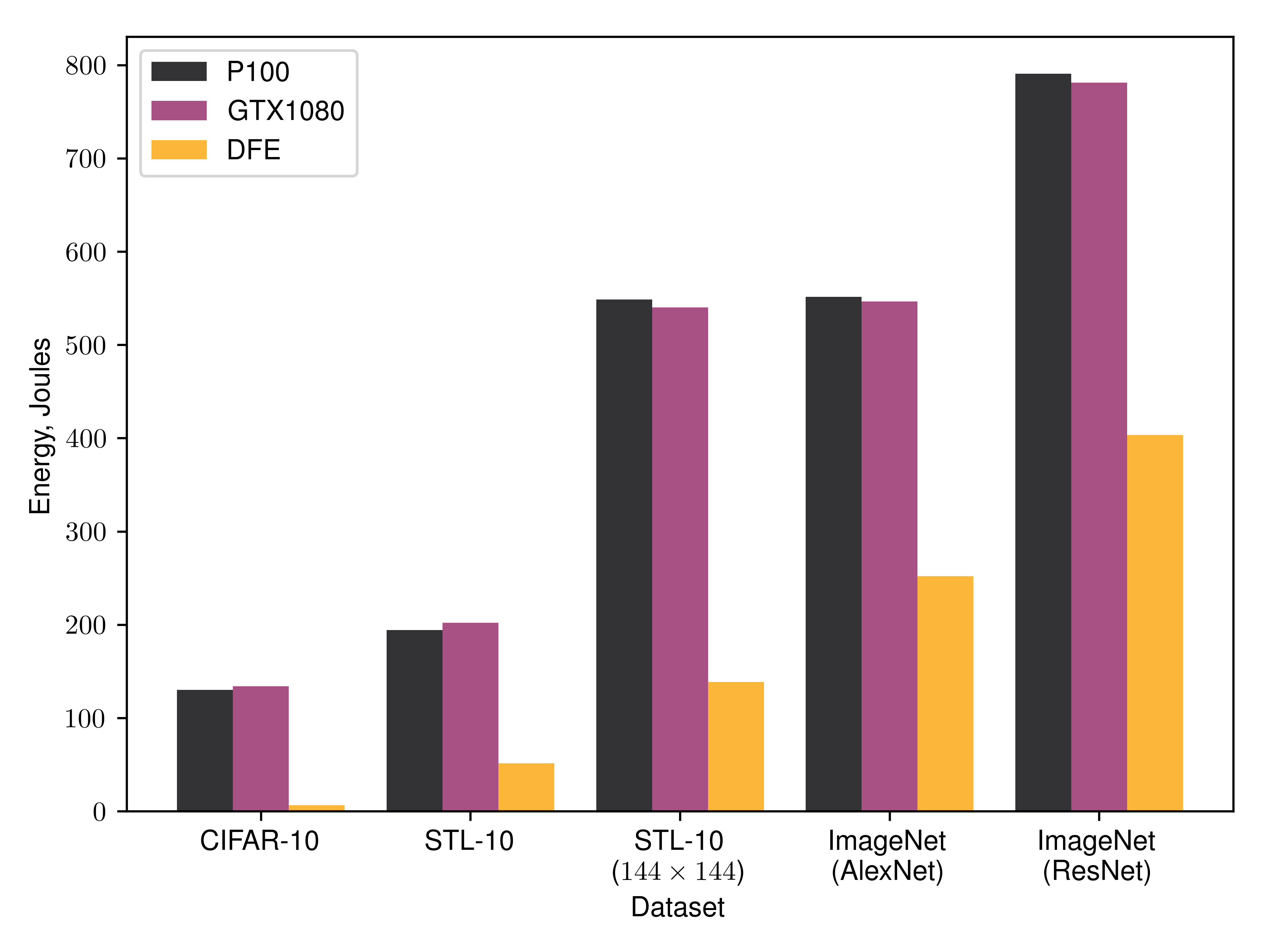}
		\vspace{-2em}
		\caption{Energy consumption comparison of FPGA- and GPU-based systems (Joules)}
		\label{energy_bars}
		\vspace{-1em}
	\end{figure}

	As shown in Figure \ref{time_bars}, for an input size of $32\times 32$, our network is $12\%$ faster than the same network running on a GPU. This presumably results from the overhead of kernel invocation processes between the CPU and GPU. Even though the GPUs demonstrate faster inference for larger inputs, power consumption of the DFE is significantly lower (at least $15\times$) for VGG-like networks, as can be seen from Figure \ref{power_bars}. For AlexNet (input size $224\times224$), the power consumption of the DFE increases, since three DFEs are needed to fit the network. 
	The energy consumption of a single-picture inference, as shown in Figure \ref{energy_bars}, is up to $20\times$ better for FPGAs, and even when more than one FPGA is used, the energy consumption was at least 50\% less compared to GPUs.

	Nevertheless, it should be noted that GPUs, unlike our architecture, are capable of simultaneously processing multiple inputs (minibatches). Modern GPUs can process at least 128-256 inputs with very small inference time degradation. While this is not helpful in real-time applications, it can speed up the process if a large amount of already-available data must be processed.
	
	\begin{figure}
		\centering	
		\includegraphics[width=0.85\linewidth]{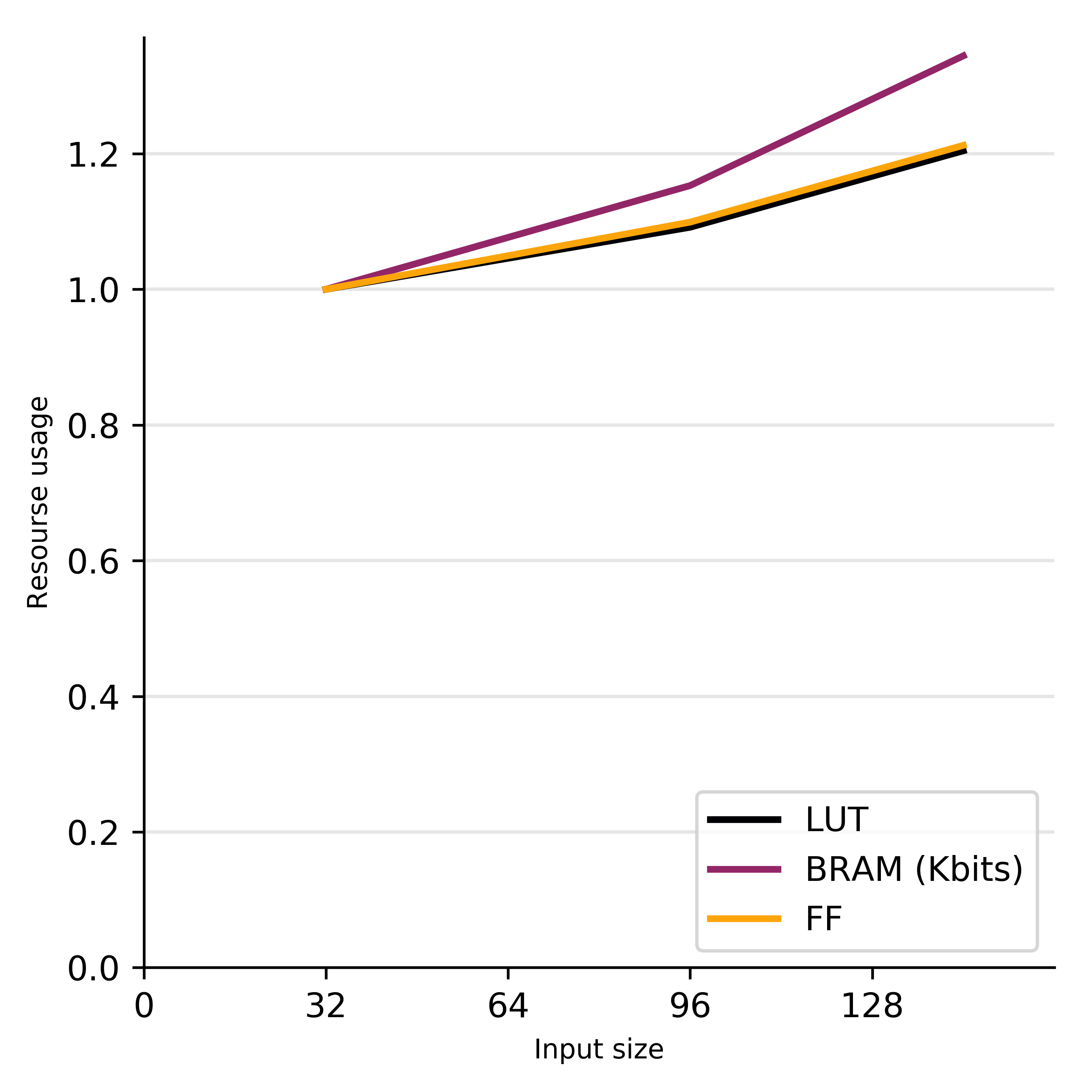}
		\vspace{-1em}
		\caption{Comparison of resource utilization for different input sizes. Change from baseline ($32 \times 32$) is shown.}
		\label{comparison_plots}
		\vspace{-1em}
	\end{figure}
	
	\subsubsection{ResNet-18 and AlexNet performance comparison}
	
	\begin{table}
		\centering
		\ra{1.3}
		
		\begin{adjustbox}{width=0.55\linewidth}
			\begin{tabular}{@{}lrr@{}} \toprule
				&AlexNet&ResNet-18\\ \midrule
				LUT&343295&596081\\
				BRAM (Kbits)&34600&30854\\
				FF&664767&1175373\\
				Run time (ms)&13.7&16.1\\
				\bottomrule
			\end{tabular}
		\end{adjustbox}
		\vspace{-.5em}
		\caption{Comparison of ResNet-18 and AlexNet networks}
		\label{AN_RN_compare}
		\vspace{-.7em}
	\end{table}
	
	To analyze the effect of adding skip connections and increasing network depth, we compared the performance of AlexNet and ResNet on DFE.
	
	First of all, it should be noted, that GPU results suggest a strong dependency on the number of layers. Since each layer waits until the previous one finishes, twice as many layers would take twice more time, even if GPU resources are not fully utilized. Our architecture, however, takes advantage of the higher amount of layers by increasing the overlapping of calculations. On a DFE, ResNet-18 takes only $17.5\%$ more time for inference, while for GPUs this number is $42.5\%$.
	
	As for resource utilization, as shown in Table \ref{AN_RN_compare}, ResNet-18 requires $\sim 75\%$ more LUTs, which is the reason we were forced to divide it into three DFEs. Due to lack of big FC layers and lower total number of parameters, ResNet requires fewer BRAMs than AlexNet. 
	
	\subsubsection{Performance comparison with other FPGA-based implementations}
	\label{Fpga_per}
	
	We compared our implementation with FINN by Umuroglu et al.\ \cite{Umuroglu:2017:FFF:3020078.3021744} using the same network architecture and dataset as appears in their paper. Their implementation, however, uses binary activations. Although the binary activations demand fewer resources and allows faster inference, multi-bit activations have superior classification accuracy \cite{DBLP:journals/corr/ZhouNZWWZ16}.  In addition, Umuroglu et al.\ store inputs in on-chip memory, while we stream them directly from the CPU. The comparison of resource utilization of both architectures is shown in Table \ref{FINN_compare_res}. Note that the resource utilization cannot be compared directly, since our implementations use FPGAs from different vendors, but we can refer to the general trends as presented. 
	
	As can be seen in the Table \ref{FINN_compare_time}, we achieve $4.1\%$ better accuracy compared to FINN, although execution and power consumption are better in their solution. We assume that a major part of the differences in runtime are due to the quality of the compilers and the special optimizations that were implemented there. Nevertheless, the main purpose of our design was to show the scalability of our solution, so less effort was directed to optimizations for small inputs.
	
	\begin{table}
		\centering
		\ra{1.3}
		
		\begin{subfigure}{0.45\linewidth}
			
			\begin{adjustbox}{width=\linewidth}
				\begin{tabular}{@{}lrr@{}} \toprule
					&FINN &DFE  \\  \midrule
					Time &0.0456 &0.8 \\
					Power &3.6  &12 \\
					Accuracy  &$80.1\%$  &$84.2\%$ \\  
					\bottomrule
				\end{tabular}
			\end{adjustbox}
			\caption{ Power (W), time (ms) and accuracy of different FPGA implementations}
			\label{FINN_compare_time}
		\end{subfigure}
		\hspace*{\fill} 
		\begin{subfigure}{0.45\linewidth}
			
			\begin{adjustbox}{width=\linewidth}
				\begin{tabular}{@{}lrr@{}} \toprule
					&FINN&DFE\\ \midrule
					LUT&46253&133887\\
					BRAM (Kbits)&6696&11020\\
					FF&-&278501\\
					\bottomrule
				\end{tabular}
			\end{adjustbox}
			\caption{Resource consumption comparison}
			\label{FINN_compare}
		\end{subfigure}
		\vspace{-.5em}
		\caption{Comparison with FINN for $32\times 32$ input size}
		\label{FINN_compare_res}
		\vspace{-.5em}
	\end{table}
	
	\subsubsection{Scalability of proposed architecture}
	\label{Scalability}
	Figure \ref{comparison_plots} shows the resource utilization of VGG-like architecture with different input sizes. It indicates that our architecture does have high scalability and the ability to effectively utilize resources on both single and multiple FPGAs. For example, increasing the size of input from $32\times32$ to $96\times96$ increases the resource utilization by approximately $5\%$ for all types of resources.
	
	Our theoretical estimation of the number of clocks per picture for ResNet-18 (the largest network implemented) is approximately $1.85 \times 10^6$. This estimation matches the measured time on a real system with a clock frequency of $105$ MHz. Among other things, this allows us to approximate runtime on next-generation FPGAs. For example, Intel's upcoming Stratix 10 FPGA promises $5\times$ higher frequency, allowing us to achieve a 3-4 ms per image inference with the same ResNet architecture, and at the same time to fit even bigger networks onto a single FPGA. 
	
	\section{Conclusions} \label{sec_conclusion}
	
	In this work, we have shown streaming architecture for QNNs, which scales well for large inputs size and large NNs. For inputs up to $144 \times 144$, resource utilization is small enough to fit on a single Stratix V 5SGSD8 FPGA. In addition, since the DFE platform allows us to easily split the network into multiple FPGAs, we can implement even larger networks, such as ResNet and AlexNet.
	
	Although GPUs outperform our implementation with large inputs, the proposed architecture is still fast enough to meet real-time requirements, achieving more than 60 fps for all types of inputs. Our results showing at least $15 \times$ lower power and $4\times$ lower energy consumption (for a single FPGA) indicate that FPGAs can be a better choice for embedded systems. In addition, the run-time is only a couple of times higher compared to the top GPUs, which allows us to speculate that next-generation FPGAs could outperform GPUs in both performance and power/energy consumption.
	
	The usage of HLS tools and DFEs as a means for functional decomposition allowed us to achieve better scalability, simplify the development process and construct a complicated FPGA system with minimal resources. Such tools may enable DL researchers with virtually no hardware development experience to construct NNs in a way similar to current scripting language frameworks, making use of the key advantages of FPGAs such as dataflow parallelism and low power consumption.
	
	\section*{Acknowledgments}
	
	The research was funded by ERC StG RAPID.
	The authors thank Maxeler Technologies Ltd for providing hardware for the experiments.
	
	\bibliographystyle{IEEEtran}
	\bibliography{ref}
	
\end{document}